# Multimodal Signal Fusion for Stress Detection Using Deep Neural Networks: A Novel Approach for Converting 1D Signals to Unified 2D Images

Yasin Hasanpoor[1†], Bahram Tarvirdizadeh[1*], Khalil Alipour[1†], Mohammad Ghamari[2†]

[1*]Department of Mechatronics Engineering, University of Tehran, North Karegar Street, Tehran 13769, Iran.
[2]Department of Electrical Engineering, California Polytechnic State University, San Luis Obispo, CA 93407, USA.

*Corresponding author. E-mail: bahram@ut.ac.ir;
Contributing authors: yasin.hasanpoor@ut.ac.ir; k.alipour@ut.ac.ir; mghamari@calpoly.edu;
[†]These authors contributed equally to this work.

## Abstract

This study introduces a novel method that transforms multimodal physiological signals—photoplethysmography (PPG), galvanic skin response (GSR), and acceleration (ACC)—into 2D image matrices to enhance stress detection using convolutional neural networks (CNNs). Unlike traditional approaches that process these signals separately or rely on fixed encodings, our technique fuses them into structured image representations that enable CNNs to capture temporal and cross-signal dependencies more effectively. This image-based transformation not only improves interpretability but also serves as a robust form of data augmentation. To further enhance generalization and model robustness, we systematically reorganize the fused signals into multiple formats, combining them in a multi-stage training pipeline. This approach significantly boosts classification performance, with test accuracy improving from 92.57% (using individual signal orderings) to 95.86% when using the combined strategy. While demonstrated here in the context of stress detection, the proposed method is broadly applicable to any domain involving multimodal physiological signals, paving the way for more accurate, personalized, and real time health monitoring through wearable technologies.

**Keywords:** photoplethysmography (PPG), galvanic skin response (GSR), multimodal signal fusion, convolutional neural network (CNN), stress recognition, 2D image matrices

## 1 Introduction

Stress is a psychological and emotional state characterized by worry and mental tension, often triggered by challenging circumstances [1]. It can be categorized as acute [2] or chronic [1], each exerting notable impacts on both physical and mental health [3]. Chronic stress, in particular, increases the risk of cardiovascular disorders [4], compromises immune function [5], and leads to mental health challenges such as anxiety [4] and depression [6]. It also impairs cognitive abilities [7], sleep [2], and social interactions [1], underscoring the importance of accurate stress detection and intervention [8–10].

Wearable devices that monitor physiological signals, such as photoplethysmography (PPG), galvanic skin response (GSR), and acceleration (ACC), have emerged as valuable tools for stress detection [11]. These devices enable real time tracking of heart rate variability [12], skin conductance [13], and physical activity [14], which are all potential indicators of stress. Integrating these signals into stress detection algorithms provides a continuous, non-invasive approach to monitoring stress in natural environments, with significant benefits for personalized healthcare [12].

Despite these advances, many existing approaches analyze individual signals in isolation [15, 16], limiting the ability to capture complex interdependencies across modalities. This separation can hinder classifier performance and generalizability, especially when deployed in real world conditions where signals may overlap or interfere [17]. Moreover, prior methods often treat multi-signal data fusion as a static preprocessing step, lacking flexibility for reconfiguration or optimization based on signal quality or task-specific importance [16–19].

Recent studies, such as Yang et al. [18], used GASF, GADF, and MTF encodings to convert 1D signals into image-based inputs, but lacked flexibility in signal selection and arrangement. Similarly, Lakshmi et al. [19] introduced a fuzzy logic and feature-fusion-based multiscale CNN (CNFSNet), yet relied on static architectures and rigid modality handling. In contrast, our method supports dynamic signal rearrangement and lightweight image fusion, making it more adaptable to diverse sensor setups and application needs.

To address these limitations and advance prior signal-to-image techniques, our work introduces a more adaptable and generalizable image transformation approach. Unlike existing fixed encodings (e.g., GASF or MTF), our method directly fuses raw signals into a configurable matrix structure. This method allows convolutional neural networks (CNNs) to learn spatial dependencies and inter-signal relationships directly from raw signal trends, without requiring predefined encoding schemes. While we applied this technique to PPG, EDA (electrodermal activity, equivalent to GSR), and ACC signals in our study, the approach is broadly applicable to any combination of multimodal signals. Additionally, by rearranging the positions of the signals (as in PEA, EAP, EPA configurations), we achieve natural data augmentation, which enhances generalization and reduces overfitting.

We evaluate the effectiveness of this approach using the WESAD dataset [12], training a CNN to classify stress and no-stress conditions. The framework converts multimodal signals into structured 2D images, enabling deep feature learning without handcrafted inputs. Section 3 details the dataset and preprocessing; Section 4 describes the image generation method; Section 5 outlines the CNN architecture; Section 6 presents evaluation and results; and Section 7 concludes the study.

## 2 Related Work and Literature Review

In recent years, stress recognition using physiological signals and deep learning has gained significant attention [20, 21]. Initial studies focused on single-signal approaches, such as PPG or GSR, to detect stress patterns (15, 22, 23). For example, [13] employed PPG to identify stress-induced changes in heart rate variability, while Navea et al. [24] and [25] used GSR to detect stress through changes in skin conductance. These studies highlight the potential of single signals for stress analysis, but may miss valuable information from other signals [20].

While these methods highlighted each signal's diagnostic potential, they lacked robustness and failed to capture inter-signal dependencies, limiting performance in real world applications [26].

### 2.1 Data Fusion and Multimodal Approaches

Recent advances in stress recognition have increasingly turned to multimodal data fusion to improve classification accuracy and robustness [17, 18, 26]. Traditional methods, as reviewed in Lazarou & Exarchos (2024), often rely on individual signals (such as PPG or GSR) and machine learning techniques like SVM, KNN, or Decision Trees [12, 20, 27, 28]. These approaches generally require extensive feature engineering and offer limited flexibility. For example, Jegan et al. [29] reported an SVM-based accuracy of 91%, which falls short compared to more recent deep learning methods [30].

To overcome such limitations, recent research has embraced deep learning architectures that can handle multiple physiological signals. For instance, Yang et al. [18] proposed a fusion strategy that combined handcrafted features (e.g., statistical measures) with deep CNN-extracted features from each signal modality, reporting a test accuracy of 90.96%. Similarly, Xiang et al. [17] applied CNNs directly to stacked raw PPG and GSR signals, treating them as multi-channel input to enable early fusion, and achieved 91% accuracy. More recently, Lakshmi et al. introduced a fuzzy-rule-based method (CNFSNet) for late fusion, combining modality-specific CNNs with adaptive weights [19]. However, such fixed architectures lack flexibility in signal arrangement and typically overlook ACC data, which is vital for handling motion artifacts. In contrast, our method introduces a dynamic image-based fusion that integrates PPG, GSR, and ACC signals into unified 2D representations, enabling flexible signal inclusion and spatial learning via CNNs.

Our work addresses these gaps through a generalizable signal fusion framework that not only integrates PPG, GSR (EDA), and ACC signals into structured 2D matrices but also supports dynamic reordering of signals (PEA, EAP, EPA configurations). This flexibility enhances both interpretability and generalization without relying on handcrafted features. Most importantly, our proposed method achieves a test accuracy of 95.86%, outperforming prior works such as those by Schmidt et al. (37). It also demonstrates scalability for broader multimodal signal classification tasks beyond stress detection.

## 2.2    Role of ACC and Emerging Deep Architecture

The accelerometer (ACC) signal plays a crucial role in identifying motion-induced artifacts in stress-related physiological data [31], yet its integration into multimodal stress detection frameworks has often been limited or underdeveloped. Tarvirdizadeh et al. (2020) explored ACC for artifact detection but did not fuse it with other modalities [32]. In 2025, Jun-Zhi Xiang et al. introduced the MMFD-SD model—a multimodal deep learning framework combining accelerometer, electrodermal, heart rate, and temperature data—showing strong real world performance (accuracy approximately 91 %) and demonstrating the value of time- and frequency-domain feature fusion [17]. However, in these approaches, the signals—although fused at the feature level—were still processed separately rather than jointly represented in a unified structure. Our method addresses this gap by incorporating ACC data directly into a unified 2D image representation alongside PPG and GSR, enabling the model to learn inter-signal relationships and improve robustness against motion artifacts.

## 2.3    CNNs and Image-Based Fusion of Biosignals

Convolutional neural networks (CNNs) have recently gained traction in physiological signal analysis, particularly for stress detection using signals such as PPG, GSR, and ACC. Hasanpoor et al. (2024) reported over 90% accuracy using a CNN–MLP model trained exclusively on PPG signals for stress detection [33]. Other studies in 2024 have incorporated wavelet-transform features with CNNs, achieving high accuracy (up to 93%) even when deployed on low-power microcontroller platforms [13, 14], underscoring their potential for real time wearable deployment. However, these methods typically process each signal independently and do not construct a unified 2D image representation that captures inter-signal dependencies across multiple modalities. This limits their ability to fully leverage the complementary nature of physiological signals for robust and generalized stress recognition.

## 2.4    Data and Dataset Reviews

Ometov et al. (2025) and Kargarandehkordi et al. (2025) reviewed open-access multimodal datasets and AI models for mental state detection, identifying limitations in synchronization, annotation, and signal fusion strategies [34, 35]. Our earlier study (Hasanpoor et al., 2022) used the UBFC dataset with a CNN-MLP model, but its lack of multimodal and motion data limited robustness[33]. In contrast, this work employs the WESAD dataset, which includes synchronized PPG, GSR, and ACC signals collected via wearable devices under semi-natural stress-inducing conditions. Its comprehensive modality coverage and inclusion of motion data make it well-suited for testing CNN-based multimodal fusion methods [12].

## 2.5    Gap Analysis & Contribution

Despite recent advances in stress detection with multimodal signals, several challenges persist:

- Most studies using multimodal signals still process each modality independently. Existing methods often rely on static fusion architectures, lacking flexibility in how modalities are combined or rearranged.
- ACC — though useful for motion artifact detection — is often excluded from fusion pipelines.
- CNN-based image fusion of physiological signals is underexplored, particularly with dynamic signal arrangements for data augmentation. Moreover, the spatial patterns created by fused multimodal signals have not been fully leveraged for deep learning–driven feature extraction.

Our work fills these gaps by:

- Fusing PPG, GSR, and ACC (or other potentially informative signals) into a unified 2D image matrix, enabling CNNs to jointly learn inter-signal dependencies within a spatially structured representation. This design allows convolutional kernels to detect localized patterns within and across modalities, which is challenging for conventional 1D approaches.
- Introducing dynamic signal-order rearrangement (e.g., PEA, EPA, EAP), serving as a novel augmentation strategy to enhance generalization and robustness.
- Achieving state-of-the-art performance (95.86% test accuracy) using a CNN trained directly on these fused representations — without the need for handcrafted features — and demonstrating improved interpretability through feature activation visualizations.

# 3    Dataset and Experimental Setup

## 3.1    Dataset Description

This study uses the WESAD dataset [12], a benchmark for stress analysis with physiological signals from 15 participants under various stress conditions, including PPG, GSR, and ACC signals essential to our stress recognition approach. The wrist-worn PPG sensor captures heart rate and pulse variability at 64 Hz, while GSR signals, sampled at 4 Hz, reflect sympathetic nervous activity. ACC signals, recorded at 32 Hz, provide wrist motion data, supporting artifact detection [12].

To ensure generalizability and avoid overfitting to individual physiological characteristics, we adopted a strict subject-independent evaluation framework, following established practices in prior works such as those by Schmidt et al. [36] and Xiang et al. [17]. Specifically, the final subject was held out entirely for testing, and the remaining subjects were split for training and validation such that all data from a given subject appeared only in one of the sets. Physiological signals were segmented into 5-second windows, and class balance (stress vs. non-stress) was preserved within each partition.

Our method encourages generalization by generating diverse signal fusion arrangements (e.g., EAP, PEA, EPA), which act similarly to data augmentation in image-based tasks. This reordering increases the representational variety seen during training, helping the model learn modality-invariant features — much like how CNNs learn to detect objects (e.g., a cat) regardless of position. Additionally, we evaluated the model in a subject-independent setup, holding out one subject entirely for testing to ensure fair cross-subject generalization assessment.

## 3.2    Data Preprocessing

The raw PPG, GSR, and ACC signals were preprocessed to improve data quality and reliability. Initial detrending removed long-term drifts and biases caused by sensor or environmental factors. Given our use of deep learning, preprocessing was minimal to allow CNNs to learn directly from raw data, avoiding excessive feature engineering [37]. This approach enabled the model to autonomously capture and distinguish stress patterns within the signals [14].

A detailed analysis of how stress physiologically affects PPG and GSR signals is provided in Supplementary Material A.

## 4    Proposed Method

This study leverages accelerometer (ACC) signals for detecting motion artifacts, following the approach in [38]. By incorporating ACC signals with PPG and GSR into the CNN architecture, the model learns to distinguish noisy signal segments. The CNN effectively identifies regions impacted by motion artifacts, allowing it to make informed processing decisions. Figure 1 shows the emergence of motion artifacts in PPG and GSR signals, marked in red. The WESAD dataset records ACC data across three directions (x, y, z), and a composite ACC signal, visualized using Equation 1, represents the combined acceleration of all three directions.

$$ACC = \sqrt{ACC_x^2 + ACC_y^2 + ACC_z^2} \qquad (1)$$

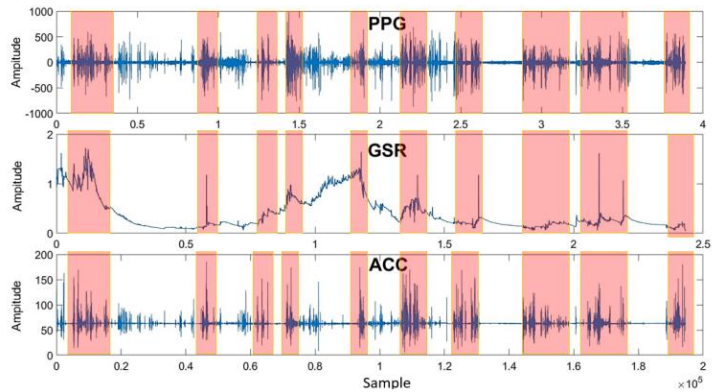

Figure 1. Visualization of motion artifacts in ACC signals and their influence on PPG and GSR signals

### 4.1   Signal Representation Matrices

In our study, we analyzed three key signals—PPG, GSR, and ACC—with sampling rates of 64 Hz, 4 Hz, and 32 Hz, respectively, transforming them into structured 2D images using a windowing approach. We divided each signal into 5-second windows, arranging the data from each window into 32 x 32 Signal Representation Matrices. As illustrated in Figure 2, these matrices visualize PPG (blue), GSR (green), and ACC (yellow) within each window.

Each 5-second window matrix contains 320 samples from PPG, 20 GSR, and 160 from ACC. This process is repeated for consecutive windows using a 1-second stride, capturing the temporal flow of the signals across successive windows.

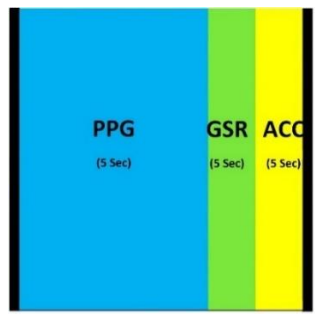

Figure 2. Distribution of PPG (blue), GSR (green), and ACC (yellow) signals within each 5-second window

The varying sampling rates posed integration challenges, addressed by sample repetition. For instance, GSR, sampled at 4 Hz, was repeated 8 times per second to balance with the other signals. Figure 3 visually details this approach, showing the arrangement and sample repetition process within the first 5-second window, with blue, green, and orange colors marking PPG, GSR, and ACC, respectively. This repetition not only aligns temporal resolution across signals to fill the image matrix but also helps to emphasize and preserve subtle features within lower-frequency signals, enhancing their visibility in the fused image representation.

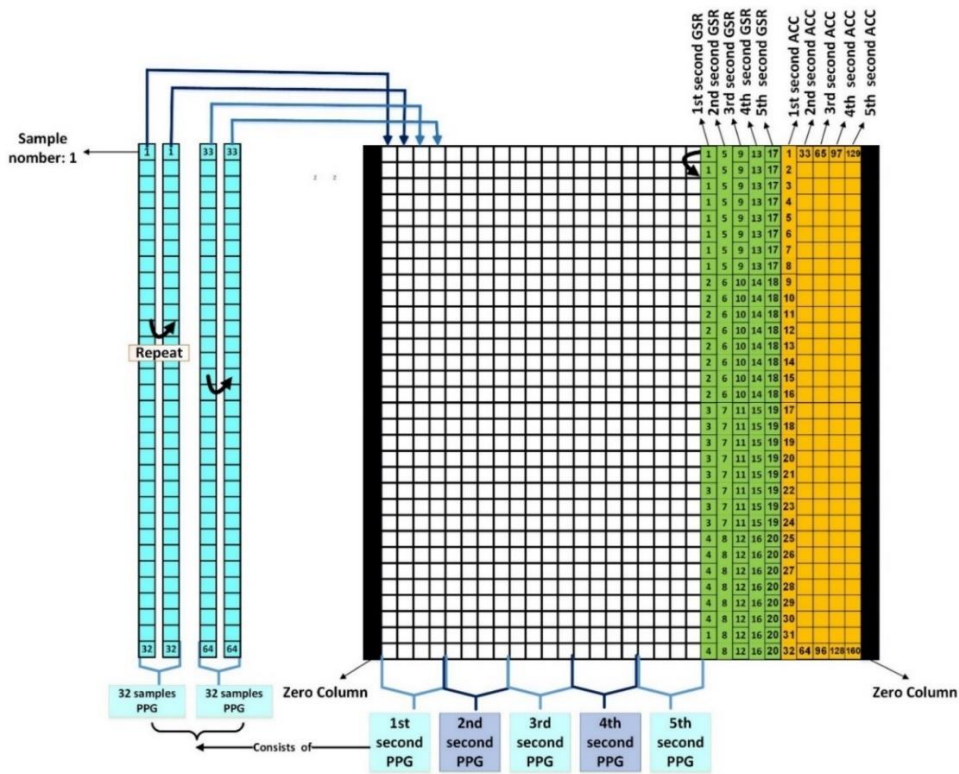

Figure 3. Matrix organization of signals in the window structure and the sample repetition process

Through this approach, we transformed the PPG, GSR, and ACC signals into structured 2D matrices, priming them for deep learning-based stress classification and analysis.

## 4.2 Image Creation Methods

In converting matrices into colored images, we explored three approaches, each impacting model accuracy differently.

The first method generated grayscale images, where each pixel's intensity corresponded to the matrix value. This approach yielded the lowest accuracy in experiments.

The second method used manual color assignments for each signal section: green for PPG, red for GSR, and blue for ACC, with color intensities based on matrix values. This slightly improved accuracy but was limited in its effectiveness.

The third and most effective method applied custom colorization, mapping matrix values to a broad color range. This technique used darker shades for lower values and brighter shades for higher ones, enhancing visual differentiation of signal magnitudes and patterns. The diverse color mapping highlighted subtle signal variations, enabling clearer pattern detection and improving stress classification accuracy. Figure 4 illustrates the color distribution for this method (range 0.00–0.95), enhancing the visibility of signal features for stress recognition.

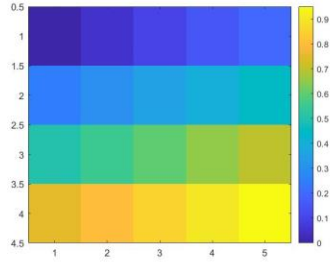

Figure 4. Distribution of colors in the custom colorization technique for signal representation matrices

By employing custom colorization, the images gained discriminative power, allowing the neural network to identify intricate signal patterns. Figure 5 compares the three methods visually, emphasizing the advantages of custom colorization for more accurate stress analysis.

In Section 6 (Results and Evaluation Metrics), we present a comparative analysis of stress recognition results across the three colorization methods. This assessment highlights the strengths of our novel data representation in advancing stress classification performance.

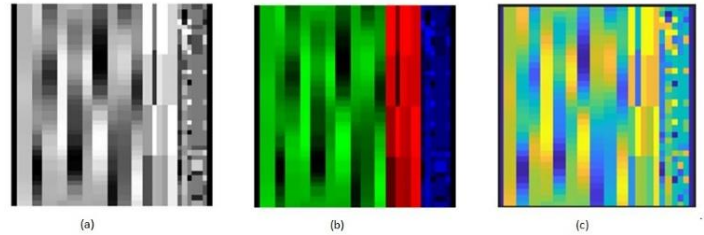

Figure 5. Comparison of colorization methods for signal representation matrices: (a) grayscale, (b) manual RGB, (c) proposed colorization techniques

## 4.3 Enhanced Neural Network Training through Dynamic Signal Arrangements

To further enhance our method, we explored alternative signal arrangements beyond the primary PEA (PPG–EDA–ACC) arrangement, specifically examining EAP (EDA–ACC–PPG) and EPA (EDA–PPG–ACC) setups. This exploration led to notable improvements in neural network learning, increasing training accuracy by 5% and test accuracy by 1%. Figure 6 compares all three arrangements across stress and no-stress classes.

From a CNN perspective, rearranging signal sequences in the fused image matrix acts as structured data augmentation. This alters the spatial location of modality-specific patterns while preserving their meaning, encouraging the model to learn modality-invariant and inter-signal relationships rather than memorizing fixed layouts—much like rotation or flipping in image augmentation. By varying the sequence, we diversify the training set and expose the model to a wider range of cross-modal dependencies, improving generalization. With n input signals, up to n! permutations are possible, expanding the training data without new collection. Additionally, the approach is scalable, allowing new physiological signals (e.g., EMG, skin temperature) to be incorporated as extra image channels to further enrich the feature space.

This versatility in signal arrangement, as demonstrated through our experiments, underscores the strength of our method in providing dynamic and enriched datasets, ultimately contributing to improved network performance and robustness in stress classification tasks.

The detailed mathematical formulation of the image construction process, including normalization, alignment, matrix assembly, and resizing, is provided in Supplementary Material B.

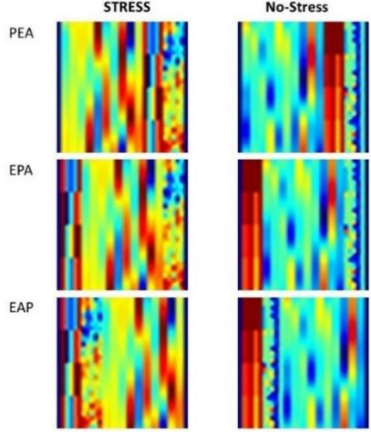

STRESS    No-Stress

PEA

EPA

EAP

Figure 6. Visual comparison of three signal arrangement strategies used for image generation: PEA (PPG + GSR + ACC), EPA (GSR + PPG + ACC), and EAP (GSR + ACC + PPG), shown from top to bottom. Each row represents a 2D image matrix derived from these arra

## 5 CNN Architecture and Design

The evaluation of our proposed multimodal signal fusion method was conducted using a CNN structure depicted in Figure 7. This architecture ensures fair comparison and reproducibility across experiments. The CNN design comprises multiple convolutional layers followed by max-pooling operations, leveraging the Rectified Linear Unit (ReLU) activation function for feature extraction and a softmax output layer for classification. The model was trained using the Adam optimizer with a learning rate of 0.001, a batch size of 64, and 16 training epochs. The detailed configuration of the CNN is elucidated below, and Figure 7 shows the configuration of the CNN classifier.

Our CNN architecture consists of the following layers:

1. **Convolutional Layers:**
   - The initial layer incorporates 64 filters with a 3×3 kernel size and ReLU activation.
   - Subsequent convolutional layers utilize 128 and 256 filters, each followed by max-pooling operations (2×2 pool size).

2. **Fully Connected Layers:**
   - Following the convolutional layers, a flattening operation is performed, and a dense layer with 128 neurons is employed, activated by ReLU.

3. **Output Layer:**
   - The final layer contains 2 neurons, representing the binary classification task, with a softmax activation function to compute class probabilities.

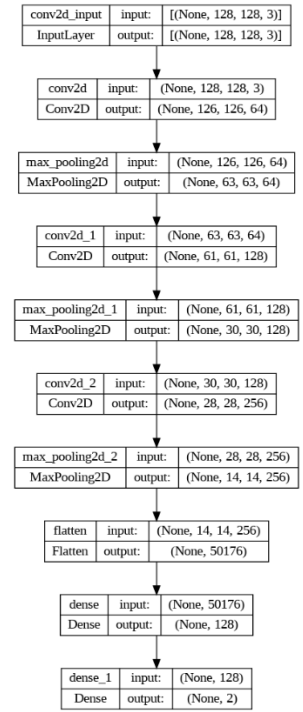

Figure 7. Configuration of CNN classifier

The CNN model is trained with the following hyperparameters:

**Image Input Size:** The input size is set to [128, 128, 3], representing the dimensions of the multimodal images.

**Train-Test Split:** We employed a subject-independent evaluation strategy by assigning one subject (S17) as the exclusive test set, while the remaining subjects (S2–S16) were used for training.

**Kernel Sizes:** Convolutional layers use 3×3 kernels for feature extraction.

**Epochs:** The training process iterates over 16 epochs to ensure convergence.

**Batch Size:** Mini-batch size is set to 64 for efficient weight updates.

**Data Shuffling:** Data is shuffled at each epoch to enhance model generalization.

**Padding:** The same padding technique is applied to preserve spatial dimensions.

**Softmax Activation:** The output layer employs the softmax activation function (Equation 2) to convert the raw outputs of the CNN into a valid probability distribution for stress classification.

$$\text{softmax}(x_i) = \frac{e^{x_i}}{\sum_j e^{x_j}} \tag{2}$$

In the softmax function, $x_i$ represents the input value for the i-th class, and the denominator normalizes all outputs so that their probabilities sum to 1. Here, K is the total number of classes, and the resulting probability distribution allows the model to assign confidence scores to each class.

**Two-Stage Dataset Arrangement and Freezing Strategy:** To enhance generalization and accuracy, we employed a two-stage training process that systematically leveraged our different signal arrangements (PEA, EPA, EAP). In Stage 1, we trained the CNN on the EAP arrangement and allowed all network parameters to update normally. After achieving convergence, we froze the feature extraction layers to preserve the learned spatial–temporal patterns. In Stage 2, we introduced additional arrangements (PEA, EPA) as new training data, updating only the classifier layers while keeping the feature extractor fixed. This approach allowed the model to retain robust low-level feature representations learned from the initial arrangement while expanding its ability to recognize complementary patterns from other arrangements, leading to improved accuracy and better cross-subject generalization.

## 6  Results and Evaluation Metrics

We evaluated four training strategies: PEA, EAP, EPA, and the combined PEA + EAP + EPA dataset. Test accuracies from Table 1 show that EAP achieved the highest single-dataset accuracy (94.43%), followed by PEA and EPA. The combined dataset reached 95.86%, demonstrating the benefit of integrating all three while preserving the strengths of EAP through our two-stage training with early-layer freezing and sample weighting.

The evaluation includes key metrics—accuracy, precision, recall, and F1 score— as well as visual insights into performance trends, confusion matrices, and ROC curves. These visual results are provided in Supplementary Material C (Figures C.1–C.3). While individual datasets yielded strong models, the combined approach proved most robust, confirming that structured multimodal integration enhances generalization without sacrificing accuracy. Table 1 summarizes performance across all strategies, showing that the two-stage training strategy with sample emphasis and feature preservation yields the best generalization and test accuracy.

Table 1 presents a comparative summary of key evaluation metrics—test accuracy, precision, recall, and F1 score— for four different training strategies: training with PEA images, EAP images, EPA images, and the combined dataset using all three (PEA + EAP + EPA). The combined approach follows a structured two-stage training strategy that preserves the strengths of EAP data while fine-tuning with selectively weighted and curated PEA and EPA samples.

The results indicate that while each individual dataset can train effective models (with EAP showing the strongest single-dataset performance), the ensemble-style strategy provides a notable increase in generalization, achieving the highest test accuracy (95.86%) and the most balanced precision, recall, and F1 scores.

This demonstrates that when diverse datasets are incorporated thoughtfully—by freezing early CNN layers, applying sample weighting, and avoiding noise overfitting—the combined training approach significantly outperforms individual dataset training. It results in a more robust classifier, capable of handling varied image styles and conditions.

Calculations for precision, recall, F1 score, and accuracy are based on the following parameters:

$$\text{precision} = \frac{TP}{TP + FP} \tag{3}$$

$$\text{recall} = \frac{TP}{TP + FN} \tag{4}$$

$$F1 - \text{score} = \frac{2 \text{xprecision} \times \text{recall}}{\text{precision} + \text{recall}} \tag{5}$$

$$\text{accuracy} = \frac{TP + TN}{TP + FN + TN + FP} \tag{6}$$

- True Positive (TP): Correctly predicted positive (no-stress class) samples.
- True Negative (TN): Correctly predicted negative (stress class) samples.
- False Positive (FP): Samples predicted as positive but actually negative.
- False Negative (FN): Samples predicted as negative but actually positive.

**Table 1**. Comparative Analysis of Performance Metrics for Different Training Approaches

| Performance Metrics | Training based on PEA images | Training based on EAP images | Training based on EPA images | Training based on PEA, EPA, and EAP images |
|---|---|---|---|---|
| Training Accuracy | 99.96% | 100.00% | 100.00% | 100.00% |
| Test Accuracy | 93.14% | 94.43% | 92.57% | 95.86% |
| Precision | 0.90 | 0.91 | 0.88 | 0.93 |
| Recall | 0.91 | 0.91 | 0.90 | 0.96 |
| F1 Score | 0.91 | 0.91 | 0.89 | 0.94 |

# 7  Discussion and Conclusion: Advancing Multimodal Signal Processing and Model Robustness

This study introduces a novel method for fusing multiple 1D physiological signals—PPG, EDA, and ACC—into unified 2D representations, enabling the use of powerful image-based deep learning models for stress classification. By encoding temporal and inter-signal relationships within a single image, the proposed method transforms traditional signal classification into a computer vision problem, significantly enhancing pattern recognition, cross-signal correlation learning, and robustness against signal noise.

The key contribution lies in this fusion-based multimodal framework, which facilitates the integration of complementary information across signals, leading to improved discriminative power. In particular, the inclusion of ACC signals was found to be essential for detecting and compensating for motion artifacts, thereby improving the model's robustness to non-physiological variability. Compared to single-modality models, this approach better preserves subtle physiological patterns that may be diluted in noise. However, the method also inherits certain drawbacks, including increased computational complexity due to higher-dimensional image representations and a potential risk of overfitting when trained on smaller datasets.

To support training stability and enhance generalization, we explored several data arrangement strategies—PEA, EAP, and EPA—and designed a two-stage learning process that selectively incorporates diverse signal arrangements. While the strongest performance was achieved with EAP-based models, integrating additional arrangements via structured fine-tuning and partial layer freezing yielded consistent, albeit modest, performance gains. This two-stage, sample-emphasis strategy aligns with ensemble learning principles, reducing overfitting risk while preserving high-level features. Table 1 summarizes performance across all strategies, showing that the proposed combination yields the best generalization and test accuracy.

Overall, the results validate our hypothesis and demonstrate that transforming multimodal biosignals into structured 2D formats not only leverages the power of CNNs but also opens new avenues for stress and affective computing.

The method's adaptability to additional modalities makes it promising for applications in continuous health monitoring, mental health assessment, and wearable technology integration. Nevertheless, broader validation on more diverse datasets, investigation into lightweight architectures for real time deployment, and improved interpretability of learned features remain important directions for future work. This balanced evaluation underscores both the promise and the current boundaries of the proposed approach, providing a clear foundation for ongoing research.

**Table 2.** Comparison of Stress Recognition Methods and Results: Single Signal vs. Multimodal Signal Fusion

| Input signal | Input Signals | Reference | Dataset | Method | Accuracy (Test) |
|---|---|---|---|---|---|
| Multiple Signal | PPG + GSR + ACC / Manual RGB | Proposed Method | WESAD | CNN | 63.71% |
| | PPG + GSR + ACC / Grayscale | | WESAD | CNN | 67.42% |
| | PPG + GSR + ACC / Custom colorization techniques | | WESAD–EAP | CNN | 94.43% |
| | PPG + GSR + ACC / Custom colorization techniques | | WESAD – PEA+EPA+EAP | CNN | 95.86% |
| | PPG + GSR | [25] | WESAD | SVM/RF | 89-93% |
| | ACC + EDA + HR | [17] | WESAD | MMFD-SD | 91% |
| | ACC + PPG + GSR + Temp + HR | [19] | Human Stress Detection Dataset | CNFSNet | 90.12% |
| | PPG + GSR + TEMP | [39] | WESAD | CNN-LSTM | 94% |
| | ECG + EDA + EEG + Video | [40] | AMIGOS | Multimodal Fusion (SVM, CNN) | 89.00% |
| | EDA + ACC + BVP + Temp | [18] | WESAD | CNN | 90.96% |
| Single Signal | PPG, GSR, ACC | [41] | Algorithmic Contest Dataset | MLP | 92.15% |
| | GSR | [42] | SWELL-KW | SVM | 82.40% |
| | GSR | [43] | WESAD | CNN-LSTM | 90.20% |
| | GSR | [24] | Custom dataset | SVM, KNN | 85.00% |
| | GSR | [44] | SGD | BayesNet | 93.73% |
| | PPG | [33] | UBFC | 1D CNN | 82.00% |
| | PPG | [45] | UBFC | LSTM | 88.44% |
| | PPG | [46] | SGD | CNN | 92.04% |
| | PPG | [47] | WESAD | CNN | 92.10% |
| | PPG, GSR, ACC, TEMP, ECG | [12] | WESAD | SVM, RF, CNN | 80-90% |

# 9  Author Contributions

Y.H. conceptualized the study, developed the methodology, conducted the experiments, and drafted the manuscript. B.T., K.A., and M.G. provided supervisory support, contributed to the design of the study, and critically revised the manuscript for important intellectual content. All authors reviewed and approved the final version of the manuscript.